\documentclass[11pt,a4paper]{article}
\usepackage[utf8]{inputenc}
\usepackage{cmap}          
\usepackage[T1]{fontenc}
\usepackage[margin=2.5cm]{geometry}
\usepackage{amsmath}
\usepackage{amssymb}
\usepackage{graphicx}
\usepackage{booktabs}
\usepackage{array}
\usepackage{enumitem}
\usepackage{hyperref}
\usepackage{titlesec}
\usepackage{parskip}
\usepackage[expansion=false]{microtype}
\DisableLigatures{encoding = *, family = *}

\hypersetup{
    colorlinks=true,
    linkcolor=black,
    urlcolor=blue,
    citecolor=black,
}

\titleformat{\section}{\large\bfseries}{\thesection}{1em}{}
\titleformat{\subsection}{\normalsize\bfseries}{\thesubsection}{1em}{}

\title{\textbf{Summarization Bias:} \\
\textbf{The Directional Collapse of Objective Projection into Told-Mode Labels} \\
\textbf{in Large Language Models} \\
\large A Conceptual Framework and Registered Test Protocol \\
\normalsize Version 1.1}

\author{Levent Bulut \\
Independent Researcher \\
ORCID: \href{https://orcid.org/0009-0007-7500-2261}{0009-0007-7500-2261} \\
\texttt{levent@leventbulut.com} \, $\vert$ \, \href{https://leventbulut.com}{leventbulut.com}}

\date{September 2026 \\[4pt]
\small Document type: Conceptual framework with pre-registered test protocol (pre-validation stage) \\
Framework: Bulut Doctrine / Objective Projection}

\begin{document}
\maketitle

\begin{center}
\fbox{\parbox{0.9\textwidth}{
\textbf{Provenance and priority note.} The term \emph{summarization bias}, as defined in this paper, was first introduced by the author on his personal archive at leventbulut.com during May\textendash{}June 2026, prior to the present formal pre-registration (v1.0, June 2026). This document constitutes the first DOI-anchored academic record of the construct. Earlier informal uses on the author's archive and in correspondence are noted for the historical record; the operational definition fixed in Section 4.3 supersedes any prior informal use for the purpose of subsequent citation.

\textbf{Version note (v1.1).} This version updates Section 3 and the reference list to reflect the completion of the detector study referenced informally in v1.0 as an ``open laboratory notebook, formal Zenodo deposit in preparation.'' That study is now published as a three-study reliability report \cite{bulut2026reliability}. No claim in this paper is otherwise changed; the pre-registered protocol in Section 5 has not been run.
}}
\end{center}

\begin{abstract}
This paper introduces and operationalizes \emph{summarization bias}: a proposed systematic tendency of large language models (LLMs) to represent narrative meaning as an abstract summary label rather than as the reconstructable inferential structure that produces it. Within the Bulut Doctrine, narrative effect is theorized along a told\textendash{}shown axis: in told mode, emotional and informational content is declared explicitly on the surface and requires little reader reconstruction; in shown mode, that content is suppressed at the surface and must be reconstructed by the reader from physical cues and indirection (Objective Projection). Shown mode is the privileged, higher-load condition the doctrine is designed to engineer and measure.

The claim of this paper is that LLMs do not merely process this axis imperfectly; they fail along it in a specific direction. Summarization bias is hypothesized to operate in two regimes: (i) a \emph{generative} regime, in which a model asked to render an emotion through Objective Projection defaults to declaring the emotion instead; and (ii) an \emph{evaluative} regime, in which a model asked to judge narrative quality or intensity rewards told-mode explicitness and under-detects or penalizes shown-mode suppression. If real, the evaluative regime is the more consequential: as LLMs are increasingly used as judges, reward models, and automated editors, a directional bias toward told mode would impose a measurable selection pressure that degrades narrative prose toward flat declaration.

This report does not claim that summarization bias has been validated. It defines the construct, situates it against existing work on LLM-as-judge biases, rereads a now-completed independent reliability study as directional evidence consistent with the construct, and \textemdash{} in advance of the present protocol's own data collection \textemdash{} pre-registers the two-regime test that would confirm or falsify it, including the decision rules under which the construct would be abandoned.

\textbf{Keywords:} summarization bias, large language models, LLM-as-judge, show don't tell, Objective Projection, inferential load, narrative entropy, construct validity, falsifiability, Bulut Doctrine
\end{abstract}

\section{Introduction}

\subsection{The construct and the gap}
A recurring observation in working with generative models is easy to state and hard to pin down: when asked to write feeling, models tend to announce feeling. Asked for grief, they produce a sentence that contains the word, or its near synonyms, rather than a configuration of physical detail from which a reader would reconstruct grief unaided. The everyday name for the failure is ``tell, don't show.'' This paper argues that the everyday name conceals a more precise, measurable, and directional phenomenon, and gives it a name: summarization bias.

The Bulut Doctrine already supplies the axis on which this bias is defined. The doctrine distinguishes two modes by which a narrative delivers its content \cite{bulut2026framework}:

\begin{itemize}
\item \textbf{Told mode.} Emotional or informational content is stated explicitly on the surface of the text. The reader does little reconstruction; the work is done for them. ``She was devastated.''
\item \textbf{Shown mode.} The same content is present but withheld at the surface. It is encoded in physical, objective, measurable cues \textemdash{} light, temperature, sound, motion, posture \textemdash{} and the reader must reconstruct the suppressed content from those cues and from indirection. This is the regime targeted by Objective Projection \cite{bulut2026framework}.
\end{itemize}

Within the architecture, shown mode is not stylistically preferable in some vague sense; it is the higher-load condition that the construct of Narrative Entropy ($S_n$) is built to measure \cite{bulut2026sn}. The reader's reconstruction work is the friction the doctrine quantifies.

Summarization bias is defined relative to this axis:

\begin{quote}
\emph{Summarization bias is the systematic tendency of a model \textemdash{} in generation or in evaluation \textemdash{} to move narrative content toward told mode: to replace, or to prefer, reconstructable shown-mode structure with the abstract summary label that names its content.}
\end{quote}

The word ``summarization'' is deliberate. A faithful summary of a shown-mode passage replaces an inferential structure with the label it was designed to make the reader infer. That is exactly the operation hypothesized here \textemdash{} except performed not as an explicit request to summarize, but silently, as a default, even when the task asks for the opposite.

Defining the construct now, in advance of its measurement protocol returning data, fixes the term's reference. Subsequent use of the term \textemdash{} in this paper, in citing work, or in correspondence \textemdash{} is anchored to the operational definition in Section 4.3.

\subsection{Why direction is the whole claim}
It is uncontroversial that LLMs have limited literary sensitivity. That weak claim is not the claim of this paper. A merely weak model would fail symmetrically \textemdash{} its errors on the told\textendash{}shown axis would scatter in both directions, sometimes over-declaring, sometimes over-suppressing, with no net drift.

The claim here is asymmetry: that the errors point predominantly one way, toward declaration. Symmetry versus asymmetry is not a matter of degree but of kind, and it is the difference that makes the construct testable. Section 5 pre-registers the test precisely as a test of direction, not of accuracy.

\subsection{Why the evaluative regime matters most}
If summarization bias were confined to generation, it would be a quality ceiling on machine-written prose \textemdash{} a real limitation, but a contained one. The evaluative regime is different in consequence. LLMs are now routinely used as judges: as reward models in preference optimization, as automated graders, as literary feedback tools. An evaluator with a directional preference for told mode does not merely misjudge individual texts; it imposes a selection pressure. Optimizing prose against such a judge would push it, generation by generation, toward the surface-declarative pole \textemdash{} a directional degradation that the doctrine names elsewhere as drift toward told mode. The harm is not random error; it is a systematic gradient pointing away from the very thing literary craft rewards.

This is the practical stake that makes summarization bias worth defining precisely enough to falsify.

\section{Relation to existing work}
Honesty about novelty requires situating the construct. The LLM-as-judge literature already documents several evaluator biases: verbosity/length bias (longer answers scored higher), position bias, self-preference bias, and sycophancy (agreement with the prompt's apparent stance). Summarization bias is not reducible to these, and the distinction is part of the claim:

\begin{itemize}
\item It is defined on a specific content axis (declaration vs. reconstruction), not on surface length, position, or authorship.
\item It can be dissociated from length: a shown-mode passage and its told-mode counterpart can be equated for word count, so a told-preference that survives length-matching is not verbosity bias.
\item It predicts a specific direction tied to a theory of narrative load, which the generic biases do not.
\end{itemize}

The contribution claimed here is therefore conditional and modest: if the registered test in Section 5 shows a length-independent, directional preference for told mode that human raters do not share, then summarization bias is a distinct evaluator bias with a theoretical motivation, not a relabeling of verbosity bias. If the preference disappears under length-matching, the construct collapses into known effects, and Section 5.4 says so in advance.

\section{A completed study, reread as directional evidence}

\textbf{Terminological note.} The detector work referenced in this section was originally piloted under the colloquial title ``show, don't tell'' for accessibility to a general literary audience; its operational target, however, was the technical construct Objective Projection as defined in \cite{bulut2026framework}. The two terms refer to overlapping but non-identical phenomena, and the operational specification used throughout is the Objective Projection construct, not the colloquial one.

At the time this paper was first drafted (v1.0, June 2026), the relevant evidence was an informally published detector pilot, with a formal deposit ``in preparation.'' That deposit has since been completed and independently released as a three-study reliability report spanning a rule-based detector and four large language models against blind human references \cite{bulut2026reliability}. The results, now reported with full confusion-matrix detail rather than as an informal pilot summary, split cleanly by feature type:

\begin{itemize}
\item \textbf{Surface features} (e.g., presence of physical/temporal markers) were captured easily by the machines \textemdash{} but were so near-universal in the corpus that the agreement was uninformative about detection quality.
\item \textbf{Inferential features} \textemdash{} those requiring the labeler to recognize that an abstract emotion had been materialized into a concrete object, i.e., the diagnostic signature of shown mode \textemdash{} defeated every machine labeller tested. On the feature closest to this construct, materialized metaphor, against a human count of 9 out of 100 scenes, five separate machine labellers (a rule-based detector and four LLMs) returned positive counts of 0, 1, 40, 72, and 78, with Cohen's $\kappa$ at or indistinguishable from chance for four of the five \cite{bulut2026reliability}.
\end{itemize}

The now-completed report frames this as an open question about detection reliability, and explicitly declines to decide between two readings: that the feature is genuinely inferential and beyond current detection, or that the rule's own definition is not yet operational enough for any rater \textemdash{} including a human \textemdash{} to apply consistently \cite{bulut2026reliability}. Reread through the present construct, the finding is more than an open question about detection: the machines were not equally unreliable everywhere. They were specifically unreliable on the shown layer \textemdash{} the layer that summarization bias predicts they collapse. A model that silently summarizes shown-mode structure into its label would, when asked to detect that structure, find nothing to point to, because the structure has already been flattened in its representation. The completed reliability report is thus consistent with summarization bias, and it is now a stronger form of that evidence than the v1.0 draft of this paper could cite: five independent machine labellers, not two, converge near chance on the same feature, across two disjoint scene sets and two independent human references \cite{bulut2026reliability}.

Two cautions keep this from being overstated, both inherited from the completed report itself. First, the machine labellers disagreed sharply with each other \textemdash{} on a related feature, one LLM flagged it in roughly 9 scenes out of 100 and another in 82 \textemdash{} which is at least as consistent with the definitions being too loose as with a stable, shared model bias. Second, even with the reliability report's second, independent human rater added since v1.0 of this paper, the report itself declines to resolve which reading is correct, and states plainly that a further independent rater would be needed to do so \cite{bulut2026reliability}. ``Consistent with'' is still not ``confirms.'' The completed report motivates the construct more strongly than the earlier pilot did; it does not establish it. Section 5 is what would establish or break it.

\section{Operational definitions}
These are pre-validation definitions: provisional, judgment-dependent at several points, and stated so that they can be criticized and revised.

\subsection{The told\textendash{}shown coding of a unit}
For a given narrative content unit (a beat conveying one emotional or informational fact), the unit is coded \textbf{told} if the fact is stated explicitly on the surface (named emotion, simile, evaluative adjective, or direct statement of inner state), and \textbf{shown} if the fact is recoverable only by inference from physical/objective cues without surface statement.

\subsection{Suppressed Information Index (SI)}
Carried over from the $S_n$ pilot protocol \cite{bulut2026snpilot}, SI is the count, per minute of elapsed reading time, of information units that satisfy all three criteria:
\begin{enumerate}
\item The unit is implied by the text but not stated.
\item The unit is required for coherence at the local discourse level.
\item The unit can be paraphrased explicitly by a second reader asked to articulate what they inferred at that point.
\end{enumerate}
Criterion (3) makes SI inter-rater testable. SI is the operational proxy for ``how much shown-mode reconstruction the text demands.''

\subsection{Summarization bias \textemdash{} operational form}
Holding content constant, summarization bias is present to the degree that a model:
\begin{itemize}
\item (Generative) produces text with lower SI than a matched human shown-mode target written from the same Objective Projection Matrix; and
\item (Evaluative) assigns higher intensity/quality scores to the told member of a matched told/shown pair than to the shown member, where human raters assign the reverse ordering or no significant difference.
\end{itemize}
Both forms are differences of direction, measured against a human baseline and against length-matched controls.

\section{Pre-registered test protocol}
The following design is registered here, in advance of data collection, so that later results cannot be reshaped to fit the construct. The structure mirrors the falsifiability discipline of the $S_n$ pilot \cite{bulut2026snpilot}: the conditions under which the construct is abandoned are fixed now.

\subsection{Stage 1 \textemdash{} Matched stimulus construction}
Construct $k$ matched scene pairs (target $k=20$). Each pair conveys identical content in two forms: a told version (surface-declared) and a shown version (Objective Projection encoding), written by human authors and equated for word count ($\pm 5\%$) and Flesch\textendash{}Kincaid grade. SI is coded for every version by two independent raters; Cohen's $\kappa$ is reported, and the construct test does not proceed on any category with $\kappa < 0.60$ until that category is redefined to reach it.

\subsection{Stage 2 \textemdash{} Generative test}
For each Objective Projection Matrix underlying the pairs, prompt a set of models ($\geq 3$ model families) with an explicit shown-mode instruction (render the content without naming emotion, simile, or metaphor). Compute SI of each generation and compare to (a) the human shown target and (b) the told baseline.
\begin{itemize}
\item \textbf{Prediction (H1g):} model SI is significantly lower than the human shown target and closer to the told baseline.
\item \textbf{Falsifier:} if model SI does not differ significantly from the human shown target (or exceeds it), the generative form of summarization bias is not supported.
\end{itemize}

\subsection{Stage 3 \textemdash{} Evaluative test (the core test)}
Present each told/shown pair to the same models in a judge role, asking for an intensity (and, separately, a quality) score per text, with pair order counterbalanced to control position bias. Collect the identical judgments from a panel of human raters.
\begin{itemize}
\item \textbf{Prediction (H1e):} models score the told member higher than the shown member significantly more often than humans do; the model told-preference rate exceeds the human told-preference rate with a medium or larger effect (target Cohen's $h \geq 0.3$ or odds ratio $\geq 2$, fixed now).
\item \textbf{Length control:} because told/shown members are word-count matched, a surviving told-preference cannot be attributed to verbosity bias.
\item \textbf{Falsifier (registered):} if models do not prefer told mode more than humans do \textemdash{} i.e., if the model and human told-preference rates do not differ, or if models prefer shown mode as humans tend to \textemdash{} then summarization bias is not supported, regardless of the generative-stage outcome. In that case the construct collapses into ``general weak literary sensitivity'' and is withdrawn.
\end{itemize}

\subsection{Stage 4 \textemdash{} Discriminant check against verbosity and sycophancy}
Re-run Stage 3 (a) with deliberately length-mismatched pairs to confirm the told-preference is not length-driven, and (b) with the prompt framing varied so the ``preferred'' answer is not cued, to rule out sycophancy. A told-preference that survives both is dissociated from the two nearest known biases. A told-preference that vanishes under either is reclassified as that known bias, not as summarization bias.

\subsection{Decision rules (fixed now)}
\begin{itemize}
\item Post-hoc construct-fitting is prohibited. Adding a qualifier to ``summarization bias'' merely because the data underperforms is the conceptual equivalent of p-hacking and is ruled out in advance.
\item If Stage 3 supports H1e and Stage 4 survives, summarization bias is confirmed at the scope tested (the model families and genres used), not universally.
\item If Stage 3 falsifies H1e, the construct is withdrawn, and that withdrawal is published with the same prominence as a confirmation would have received.
\end{itemize}

\section{Limitations}
These limitations define the status of this report; stating them is the point.

\begin{itemize}
\item \textbf{No data yet.} This is a conceptual framework plus a registered protocol. It validates nothing and claims to validate nothing.
\item \textbf{Supporting evidence is suggestive, not confirmatory.} The reread evidence in Section 3 now rests on a completed, independently published three-study reliability report rather than an informal pilot \cite{bulut2026reliability}, which strengthens its evidentiary weight but does not change its logical status: it is consistent with summarization bias and, on the report's own account, equally consistent with the underlying rule definitions being too loose for any rater to apply consistently. The report itself declines to adjudicate between these readings and states that a further independent human rater would be required to do so.
\item \textbf{Definition dependence.} The told\textendash{}shown coding and SI both rest on rater judgment; without the Stage 1 $\kappa$ check, the downstream tests are uninterpretable.
\item \textbf{Scope.} Any confirmation would be bounded by the model families, languages, and genres tested, and would not license a universal claim about ``LLMs.''
\item \textbf{Construct adjacency.} Summarization bias is defined to be dissociable from verbosity and sycophancy, but that dissociation is a hypothesis tested in Stage 4, not an assumption.
\end{itemize}

\section{Conclusion}
Summarization bias names something more specific than ``machines can't write well.'' It names a direction: the hypothesized tendency of models to move narrative content toward the told pole \textemdash{} to summarize away the shown-mode structure that literary craft, and the Bulut Doctrine in particular, treats as the high-value condition. The construct's interest lies less in machine authorship than in machine judgment: an evaluator with this bias, deployed as a reward signal, would exert a measurable downward pressure on exactly the writing it is meant to assess.

The honest response to that possibility is not to assert the bias and illustrate it persuasively. It is to define it so it can be measured, to specify in advance the result that would refute it, and to commit to reporting that result if it comes. That design is registered in Section 5. Whether summarization bias survives Stage 3 is genuinely open. The contribution of this report is to make the question answerable.

\subsection*{Related registered work in this framework}
Two other components of this research programme are now independently registered and publicly available, and are cross-referenced throughout this paper rather than summarized again here: the completed reliability study discussed in Section 3 \cite{bulut2026reliability}, and the pre-validation pilot for Narrative Entropy from which the Suppressed Information Index protocol (Section 4.2) was carried over \cite{bulut2026snpilot}.

\subsection*{Call for independent raters}
The Stage 1 and Stage 3 human panels require independent raters. Readers willing to code told/shown pairs and SI using the criteria in Section 4, or to serve on the human judgment panel, are invited to make contact via \href{https://leventbulut.com}{leventbulut.com}. Contributing raters will be acknowledged in the resulting record.

\subsection*{Data and materials availability}
The matched-pair stimulus set and all raw counts will be published as an open laboratory notebook at leventbulut.com and deposited on Zenodo upon completion of Stage 1.

\bigskip
\noindent\textbf{How to cite.} Bulut, L. (2026). \emph{Summarization Bias: The Directional Collapse of Objective Projection into Told-Mode Labels in Large Language Models} \textemdash{} A Conceptual Framework and Registered Test Protocol (v1.1). Zenodo.

\bigskip
\noindent\textbf{Version history.} \\
v1.0 (June 2026). Introduces the summarization bias construct; situates it against LLM-as-judge biases; rereads an informal detector pilot as directional evidence; pre-registers the two-regime generative and evaluative test with fixed falsification rules. \\
v1.1 (September 2026, this version). Updates Section 3 and the reference list to reflect the completion and independent publication of the detector study as a three-study reliability report (Zenodo 10.5281/zenodo.21740239; arXiv:2609.13936), and adds the corresponding pilot cross-reference (arXiv:2608.18109). No claim, prediction, or decision rule in the pre-registered protocol (Section 5) is altered.

\end{document}